%% file: custom.tex
\pdfoutput=1

\documentclass[11pt]{article}

\usepackage[final]{acl}

\usepackage{times}
\usepackage{latexsym}

\usepackage[T1]{fontenc}

\usepackage[utf8]{inputenc}

\usepackage{microtype}

\usepackage{inconsolata}

\usepackage{graphicx}
\usepackage{multirow}
\usepackage{subcaption}
\usepackage{booktabs}
\usepackage{adjustbox}
\usepackage{enumitem}
\usepackage{array, tabularx, bm}
\usepackage{arydshln}
\usepackage{xurl}
\usepackage{pifont}
\usepackage{amsmath, amsfonts}
\usepackage{tcolorbox}
\tcbuselibrary{skins,listingsutf8,breakable,poster}
\usepackage{listings}
\usepackage{courier}  
\usepackage{geometry} 
\lstdefinestyle{promptboxstyle}{
  basicstyle=\ttfamily\small,
  breaklines=true,
  columns=fullflexible,
  showstringspaces=false,
  frame=none
}

\newtcblisting{promptbox}{
  listing only,
  listing options={style=promptboxstyle},
  colback=gray!5,
  colframe=black!50,
  title={Evaluation Prompt Template},
  fonttitle=\bfseries,
  enhanced,
  sharp corners,
  boxrule=0.8pt,
  left=6pt,
  right=6pt,
  top=6pt,
  bottom=6pt
}

\setlist{nosep}

\input{preamble}

\title{Beyond Classification: Towards Speech Emotion Reasoning \\with Multitask AudioLLMs}

\author{
 \textbf{Wenyu Zhang\textsuperscript{1}},
 \textbf{Yingxu He\textsuperscript{1*}},
 \textbf{Geyu Lin\textsuperscript{1*}},
 \textbf{Zhuohan Liu\textsuperscript{1*}},
 \textbf{Shuo Sun\textsuperscript{1*}},
 \textbf{Bin Wang\textsuperscript{1*}}, \\
 \textbf{Xunlong Zou\textsuperscript{1*}},
 \textbf{Jeremy H. M. Wong\textsuperscript{1}},
 \textbf{Qiongqiong Wang\textsuperscript{1}},
 \textbf{Hardik B. Sailor\textsuperscript{1}}, \\
 \textbf{Nancy F. Chen\textsuperscript{1,2}},
 \textbf{Ai Ti Aw\textsuperscript{1}}
\\
 \textsuperscript{1}Institute for Infocomm Research (I$^\text{2}$R), Agency for Science, Technology and Research (A*STAR) \\
 \textsuperscript{2}Centre for Frontier AI Research (CFAR), Agency for Science, Technology and Research (A*STAR) \\
 Singapore
\\
 }

\begin{document}
\maketitle
\begin{abstract}
Audio Large Language Models (AudioLLMs) have achieved strong results in semantic tasks like speech recognition and translation, but remain limited in modeling paralinguistic cues such as emotion. Existing approaches often treat emotion understanding as a classification problem, offering little insight into the underlying rationale behind predictions. In this work, we explore emotion reasoning, a strategy that leverages the generative capabilities of AudioLLMs to enhance emotion recognition by producing semantically aligned, evidence-grounded explanations.
To support this in multitask AudioLLMs, we introduce a unified framework combining reasoning-augmented data supervision, dual-encoder architecture, and task-alternating training. This approach enables AudioLLMs to effectively learn different tasks while incorporating emotional reasoning.
Experiments on IEMOCAP and MELD show that our approach not only improves emotion prediction accuracy but also enhances the coherence and evidential grounding of the generated responses. Experiments on two out-of-domain datasets demonstrate the generalization capabilities of the resulting model.
\end{abstract}

\input{section/introduction}
\input{section/related_works}

\input{section/proposed_method}
\input{section/experiment_setup}

\input{section/results}
\input{section/further_analysis}
\input{section/conclusion}
\input{section/limitations}
\input{section/ethics_statement}

\section*{Acknowledgements}
This research is supported by the National Research Foundation, Singapore under its National Large Language Models Funding Initiative. Any opinions, findings, conclusions, or recommendations expressed in this material are those of the author(s) and do not reflect the views of the National Research Foundation, Singapore.

\bibliography{custom}


\appendix

\input{section/appendix}

\end{document}

%% file: preamble.tex
\newcommand{\blue}[1]{{\color{blue}#1}}
\newcommand{\orange}[1]{{\color{Orange}#1}}

\newcolumntype{P}[1]{>{\centering\arraybackslash}p{#1}}

%% file: section/introduction.tex
\section{Introduction}
\label{sec: introduction}

Recent advancements in Audio Large Language Models (AudioLLMs) \cite{he2024meralionaudiollmtechnicalreport, tang2024salmonn, chu2023qwenaudio, hu2024wavllm, Das2024SpeechVerse, Defossez2024MoshiAS} have driven significant progress in spoken language understanding, particularly for tasks focused on semantic content such as automatic speech recognition (ASR), speech translation (ST), and spoken question answering (SQA). These models typically rely on large-scale audio-text alignment to align spoken inputs with textual outputs \cite{ji2024wavchatsurvey, held2024diva}. 
However, current AudioLLMs are limited in modeling paralinguistic information, such as emotion, which is crucial for applications requiring emotionally aware or empathetic machine behavior \cite{wang2024audiobench, sakshi2025mmau, ao2024sdeval}.

Traditional emotion recognition approaches in speech primarily focus on categorical classification (e.g., predicting whether a speaker is angry or sad) \cite{emotion2vec, fu2025laerc, zhao2025steering}. While effective for high-level emotion detection, such methods offer little interpretability or reasoning about why an emotion is being expressed. .

\begin{figure*}
    \centering
    \includegraphics[width=\linewidth]{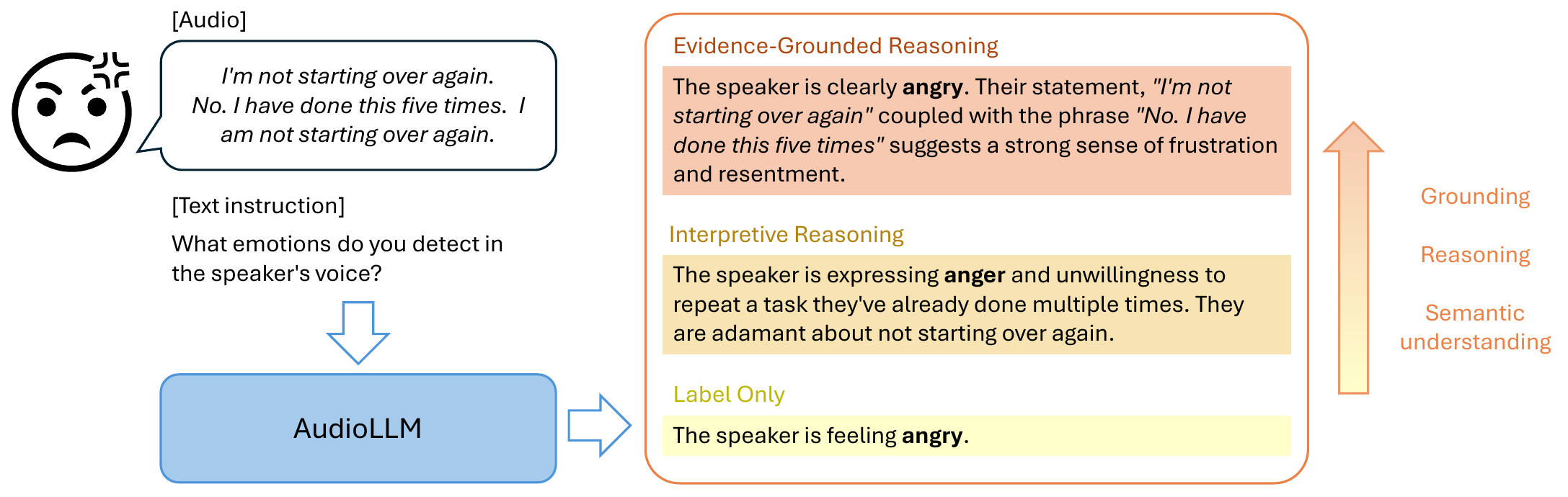}
    \caption{Overview: Our proposed method leverages the generative capabilities of AudioLLM to go beyond classification, producing emotion labels alongside grounded, transcript-informed explanations that reflect the semantic and paralinguistic content of the input speech.}
    \label{fig:overview}
    \vspace{-4mm}
\end{figure*}

In this work, we leverage the generative capabilities of AudioLLMs to incorporate reasoning \cite{ma2025audiocot, xie2025audioreasoner} as a means to improve emotion recognition. Rather than treating emotion understanding as a purely discriminative task, we guide models to generate grounded, semantically aligned explanations that reflect both what is said (semantic content) and how it is said (paralinguistic cues).
We categorize emotion recognition outputs into three types in Figure~\ref{fig:overview}:
(1) \textbf{Label Only}: direct classification (e.g. “The speaker is feeling angry”), with no explanation or grounding;
(2) \textbf{Interpretive Reasoning}: explanation via paraphrased intent or inferred state (e.g. expressing frustration due to repeated failure);
(3) \textbf{Evidence-Grounded Reasoning}: the most desirable form, which combines emotion labels with quoted utterances (e.g. “I’m not starting over again”) and interprets them to justify the emotional state.

To this end, we propose a new multitask AudioLLM framework with multi-faceted strategies across data, architecture, and training. To guide the model’s generative capabilities, we construct reasoning-augmented supervision signals from transcript-aligned data, allowing the model to learn to produce emotion explanations grounded in both linguistic and paralinguistic evidence. Our model architecture employs a dual-encoder design that disentangles semantic and emotional representations. We also propose a task-alternating training strategy that separately optimizes the semantic and emotion encoders on their respective objectives, aiming to balance performance across tasks.
Our framework is evaluated on benchmark datasets for emotion and sentiment recognition, namely IEMOCAP \cite{iemocap} and MELD \cite{poria2019meld}, as well as ASR and SQA tasks. In summary, our main contributions are:
\begin{itemize}
\item We propose a reasoning-augmented approach for speech emotion recognition, enabling AudioLLMs to generate semantically aligned, evidence-grounded explanations that enhance both interpretability and prediction accuracy.
\item We introduce a unified framework with multi-faceted strategies in data construction (reasoning target creation), architecture (dual-encoder design), and training (task-alternating training) for multitask AudioLLMs.
\item We conduct comprehensive experiments, showing that our approach effectively balances different task performances, improves emotion predictions with minimal effects on other tasks, and enables the coherence and grounding of generated emotion reasoning.
\end{itemize}

%% file: section/related_works.tex
\section{Related Works}
\label{sec: related works}

\subsection{AudioLLMs}

Multimodal large language models (LLMs), including AudioLLMs \cite{he2024meralionaudiollmtechnicalreport, tang2024salmonn, chu2023qwenaudio, deshmukh2023pengi, hu2024wavllm, Das2024SpeechVerse}, commonly adopt a modular architecture comprising three core components: (1) a modality-specific encoder that extracts features from non-textual inputs, (2) a projection or adapter module that maps these features into a representation space compatible with the LLM’s tokenizer, and (3) a pretrained LLM that generates free-form text responses based on the projected modality tokens and natural language prompts. For instance, Qwen-Audio \cite{chu2023qwenaudio} connects a Whisper-large-v2 \cite{radford2023whisper} speech encoder to the Qwen-7B \cite{bai2023qwen} language model. To capture richer audio representations, several models employ dual encoders that separately model semantic and acoustic information. SALMONN \cite{tang2024salmonn} integrates Whisper-large-v2 and BEATs \cite{chen2023beats} with Vicuna-13B \cite{vicuna2023}, while WavLLM \cite{hu2024wavllm} utilizes Whisper-large-v2 and WavLM-base \cite{chen2022wavlm}, interfaced with LLaMA-2-chat-7B \cite{Touvron2023Llama2}.

Distillation approaches use LLMs to generate responses from speech transcriptions or metadata, such as gender and emotion, to train AudioLLMs. \citet{kang2024frozen} uses an LLM to generate responses to expressive speech prompts, \citet{wang2024blsp} generate emotion-aware text continuations, and \citet{lu2024desta, lu2025desta2} generate detailed captions that reflect writing styles and tones. 

Recent works explore enabling AudioLLMs to reason. Audio-CoT \cite{ma2025audiocot} evaluates training-free chain-of-thought prompting, which requires AudioLLMs capable of general instruction following. Audio-Reasoner \cite{xie2025audioreasoner} trains with a structured reasoning framework consisting of planning, captioning, reasoning and summarizing steps. \cite{li2025rl} argues that the complex reasoning process in Audio-Reasoner may not be necessary, and the best practice remains an open research question.

\begin{figure*}
    \centering
    \includegraphics[width=\linewidth]{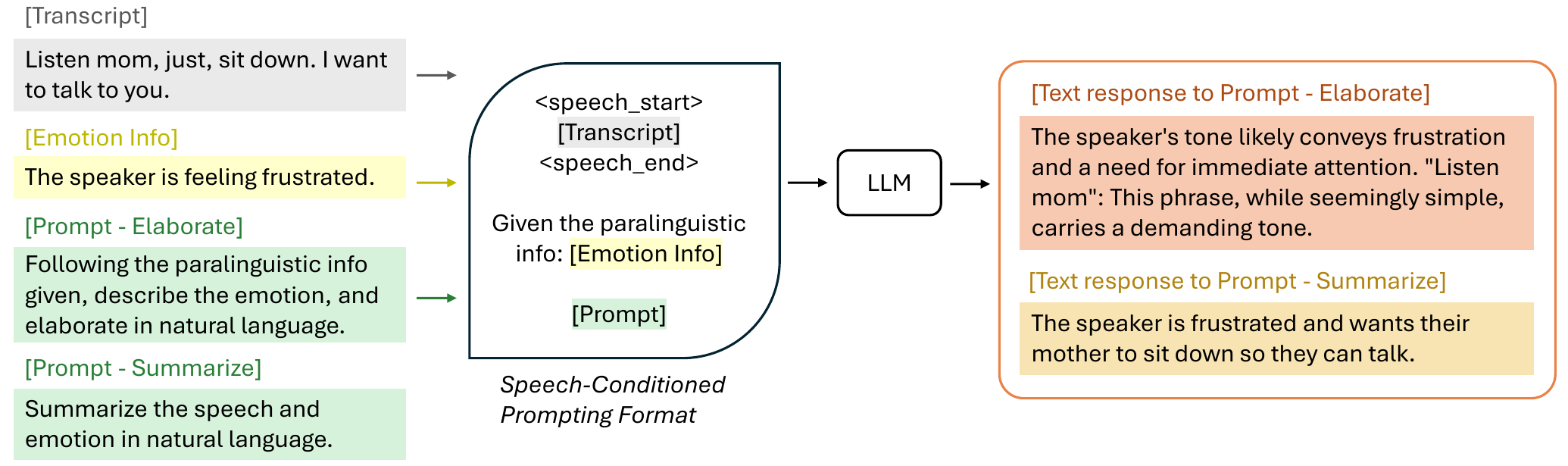}
    \caption{Emotion reason extraction: We input the transcript, emotion label, and reasoning prompt into a speech-conditioned prompting template to elicit grounded and semantically aligned emotion explanations from the LLM. The \textit{Summarize} prompt encourages interpretive reasoning based on the implied context, while the \textit{Elaborate} prompt encourages evidence-grounded reasoning based on explicit cues from the transcript. For more coarse-grained sentiment reason extraction, replace the word "emotion" with "sentiment (positive, negative, neutral)" in the reasoning prompt.}
    \label{fig:reason_extraction}
    \vspace{-4mm}
\end{figure*}

\subsection{Emotion recognition in AudioLLMs}

Recent research in emotion recognition within AudioLLMs has explored a variety of strategies to enhance affective understanding from speech \cite{bellver2024multimodal}. These approaches leverage conversational context, paralinguistic cues, and ASR-generated transcripts to improve recognition accuracy. For instance, \citet{sun2024contextual} employs ASR and LLMs in a cascaded pipeline to transcribe and analyze emotional content, though such pipelines are susceptible to error propagation. SECap \cite{xu2024secap} adopts contrastive and mutual information learning to disentangle semantic and emotional representations in speech. \citet{fu2025laerc} model speaker traits by prompting LLMs to infer emotional states based on listener responses. C²SER \cite{zhao2025steering} combines Whisper and Emotion2Vec encoders with Chain-of-Thought prompting to inject contextual reasoning into emotion classification. SpeechCueLLM \cite{Wu2024BeyondSL} introduces descriptive cues, such as volume, pitch and speaking rate, into prompts to enrich LLM inputs with prosodic information.

In contrast to prior works, which focus primarily on improving emotion classification accuracy through architectural or input-level enhancements, our approach shifts the paradigm toward emotion reasoning. Rather than outputting a single emotion label, we leverage the generative capabilities of AudioLLMs to produce semantically coherent, evidence-grounded explanations.

%% file: section/proposed_method.tex
\section{Proposed Method}
\label{sec: proposed method}

\begin{figure*}
    \centering
    \includegraphics[width=0.8\linewidth]{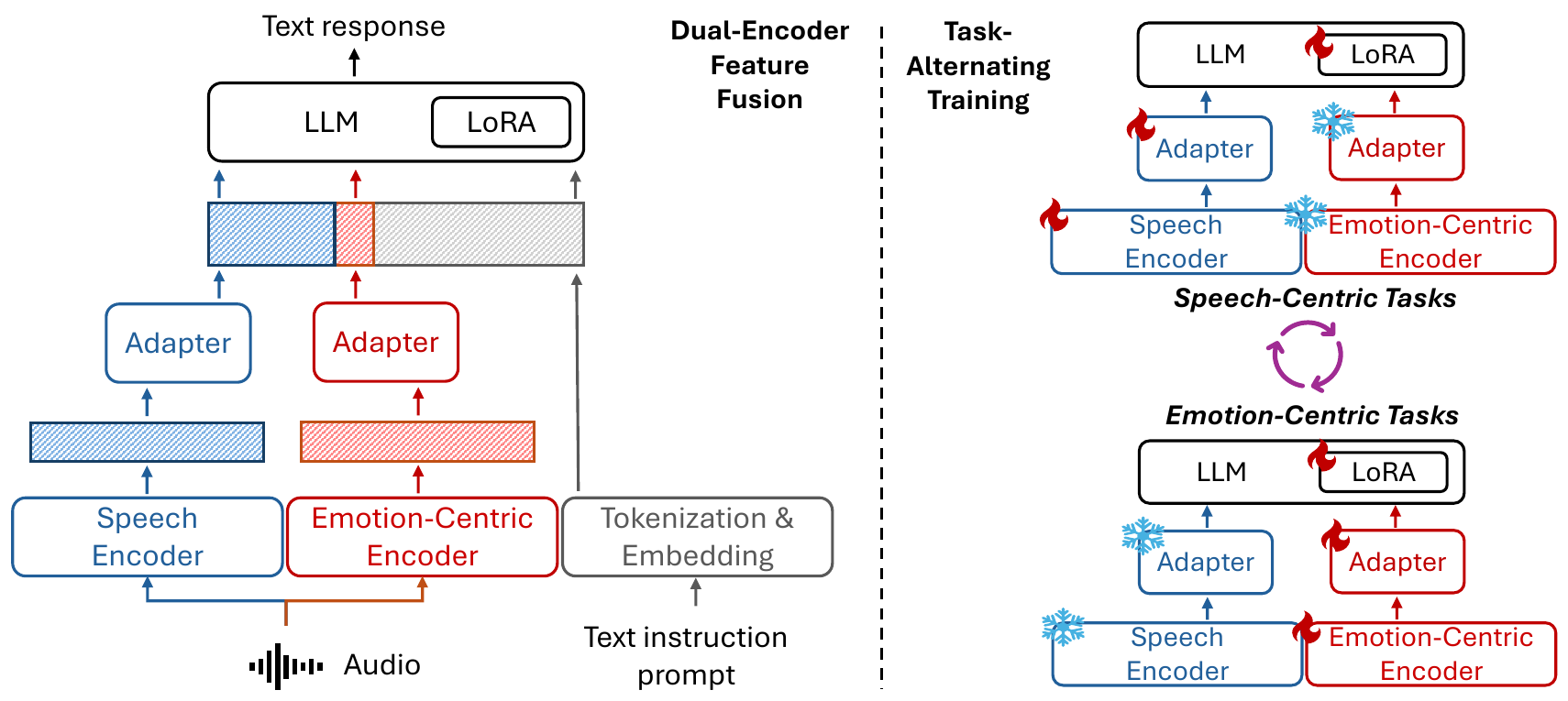}
    \caption{Dual-encoder feature fusion and task-alternating training: We combine features from a general-purpose speech encoder and an emotion-centric encoder. Each encoder and its adapter are trained by alternating between speech-centric and emotion-centric tasks, enabling effective multitask learning with disentagled representations.}
    \label{fig:training}
    \vspace{-4mm}
\end{figure*}

We propose a dual-encoder multitask AudioLLM framework that jointly models speech content and emotional reasoning. Our architecture integrates a general-purpose speech encoder and a specialized emotion-centric encoder, which are then connected to a large language model (LLM). To facilitate rich supervision, we introduce reasoning-augmented training targets derived from speech transcripts and emotion labels. Additionally, we adopt a task-alternating training strategy to ensure modular specialization and effective fusion of complementary features.

\subsection{Emotion reason extraction}

We introduce reasoning-augmented training targets that pair discrete emotion labels with natural language explanations. These explanations are derived through a prompting-based generation procedure, shown in Figure~\ref{fig:reason_extraction}. Specifically, we construct a speech-conditioned prompting format that inputs the transcript, its associated emotion label, and a reasoning prompt into an LLM. We employ two distinct prompting strategies: the \textit{Summarize} prompt encourages interpretive reasoning based on the broader implied context, while the \textit{Elaborate} prompt guides the LLM to produce evidence-grounded justifications based on explicit cues from the transcript. The resulting explanations serve as supervision signals that teach the AudioLLM to associate emotional categories with meaningful linguistic and contextual cues, such that the resulting AudioLLM attains more interpretable and context-sensitive emotion understanding. 

Using the generated targets, we construct question-answering training data by sampling questions from a curated pool designed to probe emotional understanding. These questions focus on the speaker's affective state and examples include: "How would you interpret the speaker's emotional state from their speech?", "What emotions do you think the speaker is expressing?", and "How would you describe the tone of the speaker's voice?" We apply a similar approach when querying for coarser-grained sentiment, using broader questions that elicit the speaker’s overall positive, negative, or neutral disposition.


\subsection{Dual-encoder feature fusion}

The multitask AudioLLM framework we utilize consists of: (1) a dual-encoder architecture comprising a general-purpose speech encoder \( E_{\text{speech}} \) and an emotion-centric encoder \( E_{\text{emotion}} \), each designed to capture distinct aspects of the audio input; (2) a pair of lightweight adapter modules that project the encoder outputs into a shared latent space; and (3) a pretrained LLM that consumes the fused representation to generate free-form text outputs. The emotion-centric encoder serves to enhance emotion understanding and reasoning capabilities by introducing inductive biases specific to affective cues. An overview of the dual-encoder framework is shown in Figure~\ref{fig:training}.

We denote the dataset as \((\mathcal{A}, \mathcal{T}, \mathcal{Y})\), where \(\mathcal{A}\) is the set of input audio signals, \(\mathcal{T}\) is the set of corresponding text instructions, and \(\mathcal{Y}\) is the set of output text responses. 
Given an audio input \( a_i \in \mathcal{A} \) for the \(i\)-th training sample, we extract two types of audio embeddings: the speech encoder produces \( z_{i}^{\text{speech}} = E_{\text{speech}}(a_i) \), and the emotion-centric encoder produces \( z_{i}^{\text{emotion}} = E_{\text{emotion}}(a_i) \). 
We investigated different choices of emotion-centric encoder as described in Section~\ref{sec: experiment setup} and fixed Whisper \cite{radford2023whisper}, a widely used model for automatic speech recognition (ASR), as the speech encoder throughout our experiments. The utterances are zero-padded to 30 seconds, and the encoder embeddings have sequence length 1500. 
These encoder embeddings are passed through adapter modules to be reshaped and projected into a shared latent space. The speech encoder embeddings are transformed to have a sequence length of 100, while the emotion encoder embeddings are transformed to have a shorter sequence length of 10, emphasizing condensed emotion-specific representations as a complementary signal with minimal redundancy.
We follow MERaLiON-AudioLLM \cite{he2024meralionaudiollmtechnicalreport} in our implementation of the adapter modules: we concatenate the encoder embeddings across multiple time steps to reduce the sequence length, then pass the them through a multilayer perceptron (MLP) with two hidden layers and SiLU activation. 
The resulting audio token sequences are obtained as \( \text{token}_{a_i}^{\text{speech}} \) and \( \text{token}_{a_i}^{\text{emotion}} \).
We concatenate these audio token sequences across the sequence dimension:
\[
\text{token}_{a_i} = \text{token}_{a_i}^{\text{speech}} \oplus_s \text{token}_{a_i}^{\text{emotion}}.
\]

Separately, we tokenize the text instruction \( t_i \in \mathcal{T} \) as \( \text{token}_{t_i} = \text{tokenizer}(t_i) \). The audio and text tokens are then concatenated across the sequence dimension:
\[
\text{token}_i = \text{token}_{a_i} \oplus_s \text{token}_{t_i}.
\]

Finally, the concatenated tokens are fed into the LLM to generate the target response:
\[
\hat{y}_i = \text{LLM}(\text{token}_i).
\]

\subsection{Task-alternating training}

To ensure that each encoder specializes in its respective task, we employ a task-alternating training strategy, as illustrated in Figure~\ref{fig:training}. Specifically, the speech encoder and its adapter are trained on speech-centric tasks (e.g., spoken question answering, automatic speech recognition), while the emotion-centric encoder and its adapter are trained on emotion-centric tasks (e.g., emotion recognition with explanation generation). In each training round, only the encoder corresponding to the current task, its associated adapter, and the LLM LoRA parameters are updated, while the other encoder and adapter remain frozen. This alternating scheme enables disentangled yet complementary learning of speech and emotion representations. In the final epoch, we update all adapters and the LLM LoRA parameters across all tasks to enhance multimodal alignment.


%% file: section/experiment_setup.tex
\section{Experiment Setup}
\label{sec: experiment setup}

\subsection{Model implementation}

We use Gemma-2-9B-IT \cite{Riviere2024Gemma2I} as the LLM for emotion reason extraction and in the AudioLLM framework. For each encoder, we utilize the encoder module from Whisper-Large-v3 \cite{radford2023whisper}, which is a popular choice in existing AudioLLMs \cite{tang2024salmonn, chu2023qwenaudio, hu2024wavllm, he2024meralionaudiollmtechnicalreport}. For the emotion-centric encoder, we also experiment with other options such as smaller-sized versions of Whisper, HuBERT \cite{hsu2021hubert} and Emotion2Vec \cite{emotion2vec}. For each adapter, we use a light-weight multilayer perceptron (MLP) with two hidden layers and SiLU activation function as in MERaLiON-AudioLLM \cite{he2024meralionaudiollmtechnicalreport}.
We conduct multitask training with batch size 48 for 5 epochs on 8 H100 GPUs, using an AdamW optimizer with $\beta_1=0.9$ and $\beta_2=0.999$ and a learning rate of $5\times 10^{-5}$. The prompt template for LLM input takes the form:
\begin{quote}
\textit{``\textless audio\_start\textgreater{} \{audio tokens\} \textless audio\_end\textgreater{} \{text instruction prompt\}''}
\end{quote}

\subsection{Datasets}
\label{sec: datasets}

We conduct training and evaluation on two widely used benchmarks for emotion recognition (ER) and sentiment recognition (SR): IEMOCAP \cite{iemocap} and MELD \cite{poria2019meld}. 
The IEMOCAP dataset comprises dyadic conversations between professional actors, where individual utterances are annotated with one of ten categorical emotion labels, namely anger, happiness, neutral, sadness, disgust, fear, surprise, frustration, excited and others.
MELD is a multimodal dataset derived from the TV show Friends, containing audio, video, and text for multi-party conversations. In MELD-ER, each utterance is labeled with one of seven emotion classes, namely neutral, joy, disgust, sadness, surprise, anger and fear. We also include MELD-SR for sentiment recognition, where each utterance is labeled as positive, negative, or neutral, to evaluate the model’s ability to capture overall sentiment polarity in spoken contexts.

For semantic tasks, we utilize the Spoken Question Answering (SQA) tasks in MNSC \cite{wang2025mnsc}, a corpus centered on Singlish, a Creole language rooted in English. We select MNSC because the pre-trained encoders and LLM are unlikely to have been exposed to its linguistic patterns during training, reducing the risk of performance bias from prior exposure. For further analysis in Section~\ref{sec: further analysis}, we experiment with additional SQA tasks, such as Spoken-SQuAD \cite{lee2018spokensquad} and SLUE-P2-SQA5 \cite{Shon2022SLUEPA}, and automatic speech recognition (ASR) tasks such as MNSC ASR \cite{wang2025mnsc} and LibriSpeech \cite{panayotov2015librispeech}.

Further data details and statistics are provided in the Appendix \ref{appendix: datasets}.

\subsection{Evaluation}

We perform model evaluations on datasets in Section~\ref{sec: datasets} using AudioBench \cite{wang2024audiobench} and follow its train-test splits to prevent data contamination. ASR tasks are evaluated with word error rate (WER), and remaining tasks are evaluated using LLM-as-a-Judge framework. Model outputs are assessed by Llama-3-70B-Instruct \cite{llama3modelcard} based on given scoring rubrics, and the scores are then normalized to 0-100 scale where higher scores reflect better performance.

For emotion and sentiment recognition, we grade each emotion prediction on a binary scale, where a score of 1 indicates semantic alignment with the ground-truth label. Since AudioLLMs generate open-ended responses, traditional metrics such as exact match may be insufficient. The LLM-as-a-Judge approach allows us to assess the factual correctness and relevance of model outputs in a more flexible manner. 
We analyze the evaluation approach in Section ~\ref{sec: analysis of LLM-as-a-Judge}.

We extend the LLM-as-a-Judge framework to evaluate the model's evidence-grounded emotion reasoning quality. We extract direct quotes made in the model predictions and assess them using two key metrics: Groundedness Score and Relevance Score. Groundedness assesses how well the model’s extracted quotes align with the ground truth transcript, that is, whether they are directly quoted, faithfully paraphrased, or hallucinated. Relevance measures how effectively the extracted quotes support the annotated emotion label. For each prediction, we provide the Llama-3-70B-Instruct model judge with the ground-truth emotion label, speech transcript and extracted quotes, and instruct it to assign a score from 0 to 2 for each criterion based on a given structured scoring rubric. This evaluation captures both factual alignment and emotional interpretability of the model’s output. The evaluation prompt used is provided in the Appendix \ref{appendix: evaluation}.

%% file: section/results.tex
\section{Results}
\label{sec: results}

\subsection{Effectiveness of reasoning-augmented training targets}

We evaluate the impact of different supervision targets on emotion and sentiment recognition performance using a baseline AudioLLM equipped with a single Whisper-Large-v3 encoder. From results in Table~\ref{tab: results_er_only}, we observe that models trained with reasoning-augmented targets comprising both emotion labels and natural language explanations consistently outperform those trained with label-only supervision, with almost 20\% improvement on average. This indicates that the inclusion of semantically rich and explanatory targets not only enhances interpretability but also leads to substantial gains in recognition capabilities, highlighting the effectiveness of grounding predictions in contextual reasoning. We provide examples of the different model predictions in Appendix \ref{appendix: audiollm responses}.

\input{tables/results_er_only}

\begin{figure}[tb]
    \centering
    \includegraphics[width=\linewidth]{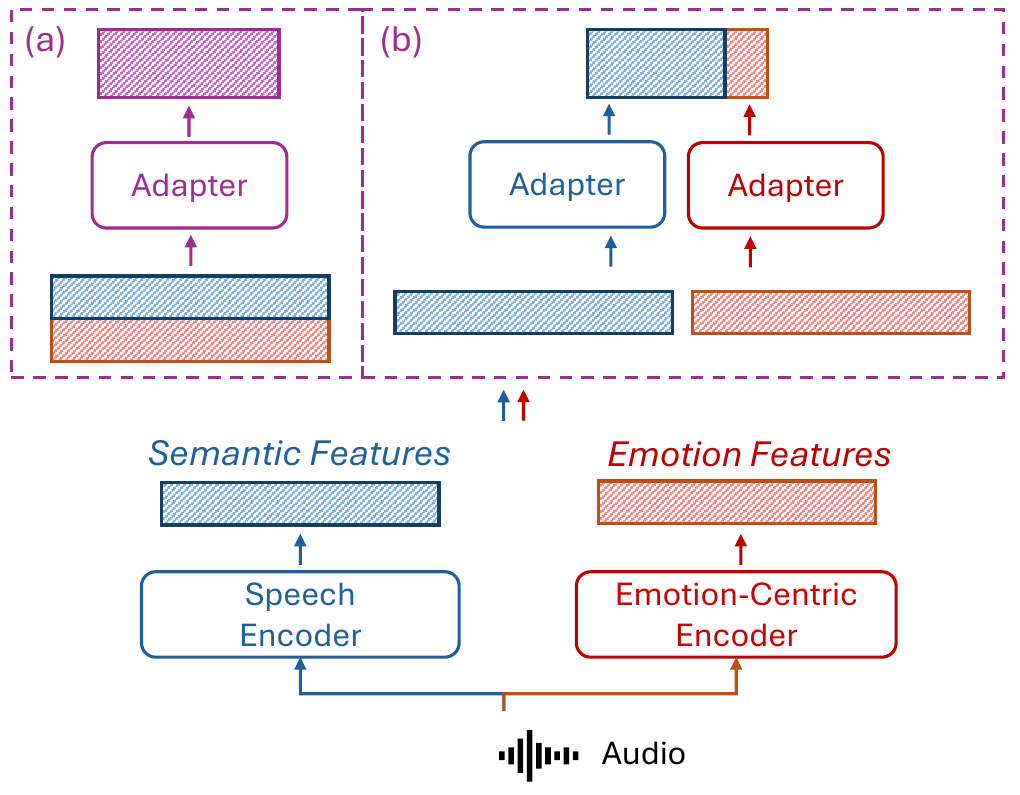}
    \caption{Strategies for feature fusion: We explore (a) fusion along the feature dimension, where features from both encoders are concatenated channel-wise, and (b) fusion along the sequence dimension, where features are concatenated token-wise across time steps.}
    \label{fig:feature_combination}
    \vspace{-4mm}
\end{figure}

\input{tables/results_feature_combination}
\input{tables/results_pl_encoder}

\subsection{Effectiveness of dual-encoder feature fusion and training}

Besides IEMOCAP and MELD, we train and evaluate on MNSC SQA Part 3-6 to assess the effects of our proposed method on non-emotion-centric tasks. Table~\ref{tab: results_feature_combination} presents a systematic comparison of key design choices, including model architecture, feature fusion strategies, and task-alternating training. We use Whisper-Large-v3 for both encoders in the dual-encoder architectures.
We observe that 
\begin{enumerate}
    \item Dual-encoder architectures can outperform single-encoder baselines, suggesting that incorporating complementary representations enhances overall performance;
    \item Concatenation along the sequence dimension yields slightly better results than concatenation along the feature dimension, likely due to better preservation of temporal structure. In the latter, both speech and emotion embeddings are reshaped to length 100, concatenated along the feature dimension, and then passed through a single adapter module. The two types of concatenation are illustrated in Figure~\ref{fig:feature_combination};
    \item Task-alternating training leads to improved performance, particularly for emotion and sentiment recognition, compared to joint multitask training.
\end{enumerate}

In Table~\ref{tab: results_pl_encoder}, we further compare multitask performance across different choices of emotion-centric encoders. We adopt a dual-encoder architecture with features concatenated along the sequence dimension, and apply task-alternating training. Our observations include:
\begin{enumerate}
\item Each round of task-specific training must be sufficiently long to ensure model convergence;
\item Encoder selection has a notable impact on performance. Specifically, Emotion2Vec+ Large, which is pre-trained for emotion recognition, provides more relevant features for emotion-centric tasks, leading to improved emotion-centric and overall performance.
\end{enumerate}



\subsection{Quality of emotion reasoning responses}

Table~\ref{tab: results_reasoning} presents the evaluation of evidence-grounded reasoning in the responses generated by our AudioLLM. Across all evaluated datasets, over 49\% of the model’s responses explicitly include direct quotes from the speech content, serving as supporting evidence for reasoning or interpretation.
The model achieves high groundedness scores, averaging 82.8\%, indicating that most quotes are faithful to the original speech transcript. Relevance scores average 65.3\%, suggesting that the majority of quotes meaningfully support the ground-truth emotion label, though there remains room for improvement, particularly on the MELD dataset.


\input{tables/results_reasoning}

\subsection{Comparison with other models}

We compare the emotion and sentiment recognition performance with end-to-end AudioLLMs evaluated in AudioBench: WavLLM \cite{hu2024wavllm}, Qwen2-Audio-7B-Instruct \cite{Qwen2-Audio}, Phi-4-Multimodal-Instruct \cite{phi4}, MERaLiON-AudioLLM \cite{he2024meralionaudiollmtechnicalreport}, Qwen-Audio-Chat \cite{chu2023qwenaudio}, SALMONN \cite{tang2024salmonn}, R1-AQA \cite{li2025r1aqa} and Audio-Reasoner \cite{xie2025audioreasoner}. R1-AQA is trained with reinforcement learning for improved thinking capabilities, and Audio-Reasoner is trained for planning and reasoning. We also compare with cascaded models that process speech in sequential stages by converting audio to text using an automatic speech recognition module before feeding the transcript into a large language model: Whisper-Large-v2 with SEA-LIONv3 \cite{sealion}, and Whisper-Large-v3 with Llama-3-8B-Instruct \cite{llama3modelcard}. From Table~\ref{tab: results_audiollm_comparison}, our proposed AudioLLM-Reasoning achieves the best performance for IEMOCAP, MELD-SR and overall.

\input{tables/results_audiollm_comparison}

\subsection{Generalization to out-of-domain datasets}

We evaluate the generalizability of the proposed model on two out-of-domain emotion recognition datasets not seen at training: M3ED \cite{zhao-etal-2022-m3ed} and CPQA \cite{cpqa2025wang}. M3ED contains utterances from Mandarin Chinese TV series. CPQA is a contexual paralinguistic question-answering dataset constructed using speech data collected from top Singapore YouTube channels; we use only the emotion recognition set for our evaluation. Both datasets are annotated for seven emotion classes, namely neutral, happy, disgust, sad, surprise, anger and fear. 
We compare our model with emotion classifier Emotion2Vec+ Large, R1-AQA which is trained with reinforcement learning for improved thinking capabilities, and Audio-Reasoner which is trained for planning and reasoning. From Table~\ref{tab: results_generalization}, our proposed AudioLLM-Reasoning achieves the best performance on both M3ED and CPQA-ER. Additional class-wise comparisons with Emotion2Vec+ Large are in Appendix~\ref{appendix: comparison with emotion2vec+}.

\input{tables/results_generalization}




%% file: tables/results_er_only.tex
\begin{table}[tb]
\centering
\setlength{\tabcolsep}{3pt} 
\begin{adjustbox}{max width=\linewidth}
\begin{tabular}{lcccc}
\toprule
\textbf{Training Targets} & \textbf{IEMOCAP} & \textbf{MELD-ER} & \textbf{MELD-SR} & \textbf{Avg} \\
\midrule
Label Only (Original)   & 18.6 & 47.9 & 48.1 & 38.2 \\ 
Interpretive Reasoning  & \textbf{60.8} & 52.6 & 60.1 & 57.8 \\
Evidence-Grounded Reasoning & 58.6 & \textbf{54.1} & \textbf{61.6} & \textbf{58.1} \\
\bottomrule
\end{tabular}
\end{adjustbox}
\caption{Emotion and sentiment recognition performance of AudioLLM (with Whisper-Large-v3 encoder) trained on different supervision targets. Training with reasoning-augmented targets yields substantial performance improvements.}
\label{tab: results_er_only}
\vspace{-4mm}
\end{table}

%% file: tables/results_feature_combination.tex
\begin{table*}[tb]
\centering
\begin{adjustbox}{max width=\linewidth}
\begin{tabular}{ll*{10}{P{1.5cm}}}
\toprule
\textbf{Concat Dim} & \textbf{Training} & \multicolumn{4}{c}{\textbf{ER/SR}} & \multicolumn{5}{c}{\textbf{SQA}} \\ \cmidrule(lr){3-6} \cmidrule(lr){7-11}
& & \textbf{IEMOCAP} & \textbf{MELD-ER} & \textbf{MELD-SR} & \textbf{Avg} & \textbf{Part 3} & \textbf{Part 4} & \textbf{Part 5} & \textbf{Part 6} & \textbf{Avg} & \textbf{Overall Avg} \\
\midrule
None                        & Joint         & 43.1 & 51.9 & 61.9 & 52.3 & 49.4 & 48.4 & 57.6 & 62.4 & 54.5 & 53.5 \\ \hdashline
\multirow{3}{*}{Feature}    & Joint         & 40.4 & 51.5 & \textbf{62.6} & 51.5 & 47.2 & \textbf{50.6} & 58.6 & 63.8 & 55.1 & 53.5 \\
                            & Alt 1 epoch   & 53.7 & \textbf{54.6} & 62.0 & \textbf{56.8} & 49.2 & 50.0 & 56.4 & 63.0 & 54.7 & 55.6 \\
                            & Alt 4 epochs  & 48.5 & 52.8 & 61.4 & 54.2 & 42.4 & 40.4 & 57.0 & 62.4 & 50.6 & 52.1 \\ \hdashline
\multirow{3}{*}{Sequence}   & Joint         & 44.1 & 48.8 & 58.1 & 50.3 & 32.8 & 30.8 & 40.2 & 47.6 & 37.9 & 43.2 \\
                            & Alt 1 epoch   & \textbf{56.6} & 52.4 & 60.7 & 56.6 & 48.0 & \textbf{50.6} & \textbf{59.2} & 63.4 & 55.3 & 55.8 \\
                            & Alt 4 epochs  & 55.1 & 52.0 & 61.5 & 56.2 & \textbf{52.4} & 49.4 & 57.6 & \textbf{64.4} & \textbf{56.0} & \textbf{56.1} \\
\bottomrule
\end{tabular}
\end{adjustbox}
\caption{Performance comparison of different methods for dual-encoder feature fusion and multitask training. Concat Dim "None" indicates the single-encoder baseline. "Joint" indicates that all tasks are trained together on all sets of encoder + adapter for 5 epochs. "Alt $x$ epoch(s)" refers to alternating training of speech-centric and emotion-centric tasks on their respective encoder + adapter every $x$ epoch(s), up to 4 epochs of data, then training all tasks on all adapters at the final epoch.}
\label{tab: results_feature_combination}
\end{table*}

%% file: tables/results_pl_encoder.tex
\begin{table*}[tb]
\centering
\begin{adjustbox}{max width=\linewidth}
\begin{tabular}{P{1.5cm}l*{10}{P{1.45cm}}}
\toprule
\textbf{Training} & \textbf{PL Encoder} & \multicolumn{4}{c}{\textbf{ER/SR}} & \multicolumn{5}{c}{\textbf{SQA}} \\ \cmidrule(lr){3-6} \cmidrule(lr){7-11}
& & \textbf{IEMOCAP} & \textbf{MELD-ER} & \textbf{MELD-SR} & \textbf{Avg} & \textbf{Part 3} & \textbf{Part 4} & \textbf{Part 5} & \textbf{Part 6} & \textbf{Avg} & \textbf{Overall Avg} \\
\midrule
\multirow{6}{*}{\shortstack{Alt 1\\epoch}}    
                                & Whisper-Large (637M)      & 56.6 & 52.4 & 60.7 & 56.6 & 48.0 & 50.6 & 59.2 & 63.4 & 55.3 & 55.8 \\
                                & Whisper-Small (88M)       & 56.8 & 52.5 & 60.6 & 56.6 & 50.8 & 50.4 & 58.6 & 61.4 & 55.3 & 55.9 \\
                                & Whisper-Tiny (8M)         & 46.2 & 49.3 & 53.1 & 49.5 & 25.8 & 23.8 & 27.0 & 32.0 & 27.2 & 36.7 \\
                                & HuBERT-XL (962M)          & 51.3 & 46.1 & 51.9 & 49.8 & 26.8 & 23.6 & 27.0 & 28.6 & 26.5 & 36.5 \\
                                & Emotion2Vec+ Large (164M) & 57.1 & 50.0 & 56.6 & 54.6 & 28.2 & 22.4 & 28.2 & 32.3 & 27.8 & 39.3 \\
                                & Emotion2Vec+ base (93M)   & 63.5 & 45.9 & 55.0 & 54.8 & 24.6 & 24.2 & 29.4 & 29.6 & 27.0 & 38.9 \\ \hdashline
\multirow{6}{*}{\shortstack{Alt 4\\epochs}}   
                                & Whisper-Large (637M)      & 55.1 & 52.0 & 61.5 & 56.2 & \textbf{52.4} & 49.4 & 57.6 & \textbf{64.4} & \textbf{56.0} & 56.1 \\
                                & Whisper-Small (88M)       & 55.3 & 52.6 & \textbf{61.6} & 56.5 & 48.6 & 49.2 & \textbf{61.4} & 61.8 & 55.3 & 55.8 \\
                                & Whisper-Tiny (8M)         & 50.3 & 52.7 & 59.7 & 54.2 & 47.8 & 47.0 & 58.0 & 62.2 & 53.8 & 54.0 \\
                                & HuBERT-XL (962M)          & 44.0 & 49.3 & 59.0 & 50.8 & 49.6 & 47.8 & 59.0 & 62.2 & 54.7 & 53.0 \\
                                & Emotion2Vec+ Large (164M) & \textbf{63.8} & \textbf{53.0} & 61.1 & \textbf{59.3} & 49.2 & \textbf{51.2} & 59.0 & 62.8 & 55.6 & \textbf{57.2} \\
                                & Emotion2Vec+ base (93M)   & 56.3 & 52.0 & 60.4 & 56.2 & 46.2 & 45.8 & 54.4 & 61.0 & 51.9 & 53.7 \\
\bottomrule
\end{tabular}
\end{adjustbox}
\caption{Performance comparison of different choices of emotion-centric encoder in the dual-encoder architecture. "Alt $x$ epoch(s)" refers to alternating training of speech-centric and emotion-centric tasks on their respective encoder + adapter every $x$ epoch(s), up to 4 epochs of data, then training all tasks on all adapters at the final epoch.}
\label{tab: results_pl_encoder}
\vspace{-4mm}
\end{table*}

%% file: tables/results_reasoning.tex
\begin{table}[tb]
\centering
\setlength{\tabcolsep}{3pt} 
\begin{adjustbox}{max width=0.8\linewidth}
\begin{tabular}{lcccc}
\toprule
\textbf{Scores} & \textbf{IEMOCAP} & \textbf{MELD-ER} & \textbf{MELD-SR} & \textbf{Avg} \\
\midrule
Quotation       & 73.8 & 49.2 & 49.6 & 57.5 \\ 
Groundedness    & 90.6 & 79.2 & 78.5 & 82.8 \\
Relevance       & 71.9 & 60.0 & 64.1 & 65.3 \\
\bottomrule
\end{tabular}
\end{adjustbox}
\caption{Evaluation of evidence-grounded reasoning. Quotation Score measures the percentage of predictions containing at least one extractable quote. Groundedness and Relevance scores (0–100 scale) assess alignment with the transcript and support for the ground-truth emotion label, respectively.}
\label{tab: results_reasoning}
\vspace{-4mm}
\end{table}

%% file: tables/results_audiollm_comparison.tex
\begin{table}[tb]
\centering
\setlength{\tabcolsep}{3pt} 
\begin{adjustbox}{max width=\linewidth}
\begin{tabular}{lcccc}
\toprule
\textbf{Model} & \textbf{IEMOCAP} & \textbf{MELD-ER} & \textbf{MELD-SR} & \textbf{Avg} \\
\midrule
Audio-Reasoner      & 51.0 & \textbf{55.9} & 54.5 & 53.8 \\
WavLLM              & 59.8 & 41.6 & 51.1 & 50.8 \\
Qwen2-Audio         & 54.0 & 41.6 & 53.9 & 49.8 \\
Cascade: Whisper+SEA-LION
                    & 44.3 & 47.4 & 56.6 & 49.4 \\
R1-AQA              & 57.2 & 42.8 & 40.7 & 46.9 \\
Phi-4-Multimodal    & 41.0 & 43.5 & 51.6 & 45.4 \\
MERaLiON            & 48.5 & 36.4 & 46.2 & 43.7 \\
Cascade: Whisper+Llama3
                    & 46.7 & 36.8 & 45.6 & 43.0 \\
Qwen-Audio          & 29.4 & 50.7 & 44.9 & 41.7 \\
SALMONN             & 23.8 & 30.5 & 41.8 & 32.0 \\ \hdashline
AudioLLM-Reasoning  & \textbf{63.8} & 53.0 & \textbf{61.1} & \textbf{59.3} \\
\bottomrule
\end{tabular}
\end{adjustbox}
\caption{Emotion and sentiment recognition performance of end-to-end AudioLLMs and cascaded models.}
\label{tab: results_audiollm_comparison}
\vspace{-4mm}
\end{table}

%% file: tables/results_generalization.tex
\begin{table}[tb]
\centering
\setlength{\tabcolsep}{3pt} 
\begin{adjustbox}{max width=0.75\linewidth}
\begin{tabular}{lcccc}
\toprule
\textbf{Model} & \textbf{M3ED} & \textbf{CPQA-ER} & \textbf{Avg} \\
\midrule
Audio-Reasoner      & 45.2 & 48.5 & 46.9 \\
Emotion2Vec+ Large  & 47.9 & 37.9 & 42.9 \\
R1-AQA              & 38.4 & 43.1 & 40.8 \\ \hdashline
AudioLLM-Reasoning  & \textbf{48.6} & \textbf{49.0} & \textbf{48.8} \\
\bottomrule
\end{tabular}
\end{adjustbox}
\caption{Emotion recognition performance on out-of-domain datasets.}
\label{tab: results_generalization}
\vspace{-4mm}
\end{table}

%% file: section/further_analysis.tex
\section{Further Analysis}
\label{sec: further analysis}

We further investigate whether emotion understanding capabilities can be introduced into a model that was not originally trained for these tasks. Starting from a base AudioLLM without any emotion-specific supervision, we explore adding an emotion-centric encoder to the architecture. We train the emotion-centric encoder Emotion2Vec+ Large, adapter and LLM LoRA on emotion-centric tasks, then finetune the adapters and LLM LoRA on all tasks.

Table \ref{tab:results_injection} presents the effect of incorporating emotion supervision into a base AudioLLM trained on different upstream tasks. Across both training configurations, we observe substantial gains in emotion and sentiment recognition, with improvements of +12.3 and +22.4 points, respectively. This enhancement in emotional understanding comes with a slight degradation in the model’s performance on the original tasks. For instance, SQA performance remains comparable, and ASR performance is largely preserved or slightly improved. These results highlight that emotional capabilities can be effectively injected into a multimodal model without sacrificing its existing competencies.

\input{tables/results_injection}

\section{Assessment of LLM-as-a-Judge metric}
\label{sec: analysis of LLM-as-a-Judge}

We adopt the LLM-as-a-Judge metric for evaluation as the AudioLLMs can generate open-ended and expressive outputs. Traditional metrics relying on exact string matching are insufficient in this context. We identify three main cases where traditional metrics can fail to capture the true quality of model responses: 1. the model uses semantically equivalent but lexically different expressions to describe emotions;
2. the model output includes multiple plausible or related emotional states; and
3. the predicted labels fall into semantically overlapping categories (e.g., excited vs. happy, or anger vs. frustration). Outside of these special cases,  the LLM-as-a-Judge metric effectively reduces to an accuracy-like measure, where the model's prediction is compared against a reference by string matching.

Out of the 858 test samples in IEMOCAP, 27.6\% of the AudioLLM-Reasoning emotion predictions fall into the special cases. Out of the 2610 test samples in MELD-ER, 6\% of the AudioLLM-Reasoning emotion predictions fall into the special cases. To better understand the characteristics of the LLM-as-a-Judge metric, we randomly selected 50 special-case samples each from the IEMOCAP and MELD-ER datasets to be assessed by 4 human evaluators. The evaluators are provided with the audio clips, references and model's answers, and are instructed to rate the model's answers based on their alignment with the audio clips and references. For both IEMOCAP and MELD-ER, only 2\% of the selected samples are scored as correct by the LLM judge but scored as wrong by the human evaluators. In contrast, 16\% of the IEMOCAP selected samples and 30\% of the MELD-ER selected samples are scored as wrong by the LLM judge but scored as correct by the human evaluators, suggesting that the LLM judge tends to be more conservative or strict in its assessments compared to human evaluators.

%% file: tables/results_injection.tex
\begin{table}[tb]
\centering
\setlength{\tabcolsep}{3pt} 

\begin{subtable}[t]{\linewidth}
\centering
\begin{adjustbox}{max width=0.85\linewidth}
\begin{tabular}{lccc}
\toprule
\textbf{Model} & \textbf{ER/SR ($\uparrow$)} & \textbf{SQA ($\uparrow$)} & \textbf{ASR ($\downarrow$)}\\
\midrule
Base AudioLLM                 & 44.1 & \textbf{56.0} & \textbf{19.5}\\
\quad + Emotion Supervision   & \textbf{56.4} & 54.1 & 19.6 \\
\bottomrule
\end{tabular}
\end{adjustbox}
\caption{Base AudioLLM trained on MNSC SQA Part 3-6 and MNSC ASR Part 3-6.}
\end{subtable}

\vspace{1em}

\begin{subtable}[t]{\linewidth}
\centering
\begin{adjustbox}{max width=0.85\linewidth}
\begin{tabular}{lccc}
\toprule
\textbf{Model} & \textbf{ER/SR ($\uparrow$)} & \textbf{SQA ($\uparrow$)} & \textbf{ASR ($\downarrow$)}\\
\midrule
Base AudioLLM               & 35.6 & \textbf{80.3} & 3.8 \\
\quad + Emotion Supervision & \textbf{58.0} & 79.0 & \textbf{3.6} \\
\bottomrule
\end{tabular}
\end{adjustbox}
\caption{Base AudioLLM trained on Spoken-SQuAD and SLUE for SQA, and LibriSpeech Clean and Other splits for ASR.}
\end{subtable}

\caption{Effect of adding emotion supervision on a trained base AudioLLM. Adding emotion supervision improves emotion understanding with slight compromise on the performance of other tasks}
\label{tab:results_injection}
\end{table}

%% file: section/conclusion.tex
\section{Conclusion}
\label{sec: conclusion}

In this work, we propose a unified framework that brings emotion reasoning into multitask AudioLLMs, combining dual encoders, reasoning-augmented supervision, and task-alternating training. Our method improves emotion and sentiment recognition and enables the generation of evidence-grounded explanations, as demonstrated on IEMOCAP and MELD benchmark datasts. This work highlights the potential of generative AudioLLMs for more interpretable and emotionally aware speech understanding.

%% file: section/limitations.tex
\section{Limitations}
\label{sec: Limitations}

While our framework significantly improves both emotion recognition and explanation capabilities in AudioLLMs, several limitations remain. The explanation generation quality can vary across emotions and speaker styles, especially for subtle or ambiguous affective states. Moreover, the quality of extracted emotion reasoning is dependent on the capabilities of the teacher LLM, which may introduce biases or inaccuracies in supervision. The current benchmarks for emotional reasoning are limited in diversity and scale, highlighting the need for more comprehensive evaluation datasets that capture a wider range of emotional expressions and contextual richness.

%% file: section/ethics_statement.tex
\section{Ethics Statement}
\label{sec: ethics statement}

All datasets used in our study, including IEMOCAP and MELD, are publicly available and widely used in the research community. However, we emphasize that caution is necessary when deploying emotion-aware AI systems in real-world or sensitive contexts, as misinterpretation of emotional cues may lead to unintended consequences. Ensuring transparency, user consent, and appropriate safeguards is critical when applying these technologies beyond academic settings.

%% file: section/appendix.tex
\section{Experiment Details}
\label{appendix: experiment details}

\subsection{Datasets}
\label{appendix: datasets}

We conduct training and evaluation on two widely used benchmarks for emotion recognition (ER) and sentiment recognition (SR): IEMOCAP \cite{iemocap} and MELD \cite{poria2019meld}. IEMOCAP is made available under a custom non-commercial research license, and MELD is distributed under the GNU General Public License v3.0 (GPL-3.0). Since IEMOCAP lacks a predefined train-test split, we adopt the 90-10 split defined in AudioBench \cite{wang2024audiobench}, with 9035 training samples (anger: 1140, disgust: 2, excited: 1816, fear: 98, frustration: 2608, happiness: 588, neutral: 1539, other: 23, sad: 1120, surprise: 101) and 1004 test samples (anger: 129, disgust: 0, excited: 160, fear: 9, frustration: 309, happiness: 68, neutral: 187, other: 3, sad: 130, surprise: 9). MELD has 9988 training samples (anger: 1109, disgust: 271, fear: 268, joy: 1743, neutral: 4709, sad: 683, surprise: 1205) and 2610 test samples (anger: 345, disgust: 68, fear: 50, joy: 402, neutral: 1256, sadness: 208, surprise: 281).

For semantic tasks, we utilize the Multitask National Speech Corpus (MNSC) \cite{wang2025mnsc}, specifically SQA Part 3-6 and ASR Part 3-6, released under the Singapore Open Data License. We also use Spoken-SQuAD \cite{lee2018spokensquad} released under CC-BY-SA-4.0 License, SLUE-P2-SQA5 \cite{Shon2022SLUEPA} which is a collection of datasets released under CC-BY-SA-4.0 License and Apache License 2.0, and LibriSpeech \cite{panayotov2015librispeech} released under CC-BY-4.0 License.

All experiments in this work respect the respective licenses and usage terms of the datasets.

\subsection{Comparison with Emotion2Vec+}
\label{appendix: comparison with emotion2vec+}

We compare the performance of AudioLLM with emotion classifier Emotion2Vec+ Large. For the in-domain dataset MELD-ER in Table~\ref{tab:results_ua_wa}, AudioLLM trained with reasoning-augmented targets achieves the highest weighted average score  (53.0\%), outperforming both the label-only variant (47.1\%) and Emotion2Vec+ Large (44.7\%). As the AudioLLMs are trained on a limited set of emotion datasets (i.e. IEMOCAP and MELD), they tend to fit to the training label distribution. In contrast, Emotion2Vec+ Large is trained on five emotion datasets.

The performance on the out-of-domain dataset CPQA-ER in Table~\ref{tab:results_ua_wa_generalization} is affected less by the training label distribution. AudioLLM-Reasoning outperforms Emotion2Vec+ Large classifier on 6 out of 7 classes, and significantly outperforms the classifer on both the unweighted average score (45.3\% vs. 36.0\%) and the weighted average score (49.7\% vs. 38.3\%). Moreover, the discriminative classifier cannot readily extend beyond its label space. In Table~\ref{tab:results_ua_wa_generalization}, 11 samples involving states such as frustration, embarrassment, and mixture of emotions are excluded. The vanilla AudioLLM trained with label-only targets lacks generalization capabilities and has severely degraded performances.

\input{tables/results_ua_wa}

\input{tables/results_ua_wa_generalization}

\subsection{AudioLLM responses}
\label{appendix: audiollm responses}

We conduct a qualitative analysis of model predictions across different reasoning formats on the IEMOCAP, MELD-ER and MELD-SR datasets. Each example includes the transcript, ground-truth label, and model-generated outputs under three supervision types: label-only, interpretive reasoning, and evidence-grounded reasoning. As shown in Table~\ref{table:audiollm responses iemocap}, \ref{table:audiollm responses meld-er} and \ref{table:audiollm responses meld-sr}, label-only responses often fail to capture the correct emotional nuance, defaulting to neutral predictions even when the emotion is apparent. In contrast, interpretive and evidence-grounded reasoning better align with the ground truth, offering richer justifications and improved emotion recognition. Notably, evidence-grounded reasoning demonstrates superior clarity by explicitly linking speech content and affective cues to the predicted emotion.

\input{tables/audiollm_responses_iemocap}
\input{tables/audiollm_responses_meld_emotion}
\input{tables/audiollm_responses_meld_sentiment}

\subsection{Evaluation of emotion reasoning quality}
\label{appendix: evaluation}

We follow AudioBench's LLM-as-a-Judge framework to evaluate the model's evidence-grounded emotion reasoning quality. We extract direct quotes made in the model predictions and assess them using two key metrics: Groundedness Score and Relevance Score. For each prediction, we provide the Llama-3-70B-Instruct model judge with the ground-truth emotion label, speech transcript and extracted quotes, and instruct it to assign a score from 0 to 2 for each criterion based on a given structured scoring rubric. The scores are then normalized to 0-100 scale. The evaluation prompt used is shown in Figure~\ref{fig:evaluation-prompt}.

\begin{figure*}
\centering
\begin{promptbox}
[Ground Truth Emotion]
{reference}

[Ground Truth Transcript]
{transcript}

[Extracted Quotes from Model Prediction]
{extracted_quotes}

[Evaluation Task]
Evaluate the extracted quotes using the following three criteria.

**Groundedness Score**
Assess whether the extracted quotes are grounded in the ground truth transcript.
Scoring Guide:
Score0: The quotes do not appear in the ground truth transcript and are not semantically aligned (i.e., hallucinated or generic).
Score1: The quotes partially match the ground truth transcript. There may be loose paraphrasing or selective grounding.
Score2: The quotes are clearly derived from the ground truth transcript, through direct quotes or faithful paraphrases.

**Relevance Score**
Assess whether the extracted quotes support the ground truth emotion label.
Scoring Guide:
Score0: The quotes are irrelevant or inconsistent with the ground truth emotion. They may even suggest a different emotion.
Score1: The quotes are loosely related to the ground truth emotion but lack clarity, specificity, or completeness.
Score2: The quotes clearly and directly support the ground truth emotion.

Respond with the following structured format:

Ground Truth Emotion: (string)
Ground Truth Transcript: (string)
Extracted Quotations from Model Prediction: (list of strings)
Groundedness Score: (int)
Relevance Score: (int)
Explanation: (string - justify the assigned scores)
\end{promptbox}
\caption{Evaluation prompt used for assessing groundedness and relevance of extracted emotional evidence.}
\label{fig:evaluation-prompt}
\end{figure*}

%% file: tables/results_ua_wa.tex
\begin{table}[tb]
\centering

\begin{adjustbox}{max width=\linewidth}
\begin{tabular}{l*{4}{P{2.0cm}}}
\toprule
\textbf{Class} & \textbf{Num Samples} & \textbf{Emotion2Vec+ Large} & \textbf{AudioLLM - Label Only} & \textbf{AudioLLM - Reasoning} \\
\midrule
Neutral       & 1256 & 54.2 & 64.7 & \textbf{83.8} \\
Joy           & 402  & \textbf{54.5} & 41.5 & 36.3 \\
Disgust       & 68   & 0.0  & 8.8  & \textbf{13.2} \\
Sadness       & 208  & \textbf{32.2} & 23.1 & 15.9 \\
Surprise      & 281  & \textbf{38.8} & 28.8 & 26.0 \\
Anger         & 345  & 26.1 & \textbf{32.2} & 31.3 \\
Fear          & 50   & 2.0  & 6.0  & \textbf{8.0} \\
\midrule
Unwt Avg  & 2610 & \textbf{29.7} & 29.3 & 29.2 \\
Wt Avg    & 2610 & 44.7 & 47.1 & \textbf{53.0} \\
\bottomrule
\end{tabular}
\end{adjustbox}

\caption{In-domain comparison: Class-wise emotion recognition performance comparison of Emotion2Vec+ Large classifier vs. AudioLLM (with Emotion2Vec+ Large emotion-centric encoder) trained on Label Only or Reasoning targets on MELD-ER. The unweighted average (Unwt Avg) treats all classes equally regardless of sample size, while the weighted average (Wt Avg) reflects the sample distribution by weighting each class accordingly.}
\label{tab:results_ua_wa}
\end{table}

%% file: tables/results_ua_wa_generalization.tex
\begin{table}[tb]
\centering

\begin{adjustbox}{max width=\linewidth}
\begin{tabular}{l*{4}{P{2.0cm}}}
\toprule
\textbf{Class} & \textbf{Num Samples} & \textbf{Emotion2Vec+ Large} & \textbf{AudioLLM - Label Only} & \textbf{AudioLLM - Reasoning} \\
\midrule
Neutral       & 102 & 12.7 & 13.7 & \textbf{32.4} \\
Joy           & 91  & 71.4 & 38.5 & \textbf{74.7} \\
Disgust       & 30  & 13.3 & 10.0 & \textbf{20.0} \\
Sadness       & 57  & \textbf{59.6} & 22.8 & 40.4 \\
Surprise      & 54  & 31.5 & 1.9 & \textbf{33.3} \\
Anger         & 83  & 43.4 & 28.9 & \textbf{78.3} \\
Fear          & 50  & 20.0 & 4.0 & \textbf{38.0} \\
\midrule
Unwt Avg  & 467 & 36.0 & 17.1 & \textbf{45.3}\\
Wt Avg    & 467 & 38.3 & 19.7 & \textbf{49.7} \\
\bottomrule
\end{tabular}
\end{adjustbox}

\caption{Out-of-domain comparison: Class-wise emotion recognition performance comparison of Emotion2Vec+ Large classifier vs. AudioLLM (with Emotion2Vec+ Large emotion-centric encoder) trained on Label Only or Reasoning targets on CPQA-ER. 11 samples with ground-truth labels outside of the label space of Emotion2Vec+ Large classifier have been excluded.}
\label{tab:results_ua_wa_generalization}
\vspace{-4mm}
\end{table}

%% file: tables/audiollm_responses_iemocap.tex
\begin{table*}[t]
\centering
\small
\renewcommand{\arraystretch}{1.3}
\begin{tabular}{|p{15.5cm}|}
\hline
\textbf{\underline{Example 1}} \\
\textbf{Transcript:} \textit{\orange{Is he an actor too, or-?}} \\
\textbf{Ground Truth Emotion Label:} \blue{Neutral} \\ \hdashline
\textbf{Label Only:} The speaker's speech reflects a \blue{neutral} state \\
\textbf{Interpretive Reasoning:} The speaker is asking if someone was an actor too, implying a shared experience or connection. They seem \blue{neutral} in their tone, neither excited nor disappointed. \\
\textbf{Evidence-Grounded Reasoning:} The speaker's tone is \blue{neutral}, suggesting they are neither expressing strong positive nor negative emotions. They are simply seeking clarification or information. The phrase "Is he an actor too?" implies a previous context where someone was identified as an actor. The speaker is now inquiring if another person also shares this profession. \\
\\
\hline
\textbf{\underline{Example 2}} \\
\textbf{Transcript:} \textit{\orange{Oh, I will never forget his face. Ooh-rah [LAUGHTER].}} \\
\textbf{Ground Truth Emotion Label:} \blue{Excited} \\ \hdashline
\textbf{Label Only:} The speaker's speech suggesting a \blue{neutral} state. \\
\textbf{Interpretive Reasoning:} The speaker is laughing and reminiscing about someone's face, implying a positive and humorous memory. They express \blue{excitement and amusement}. \\
\textbf{Evidence-Grounded Reasoning:} The speaker is clearly \blue{excited and perhaps a little bit nostalgic}. Here's why: \orange{"Oh, I will never forget his face"}: This phrase suggests a strong memory, likely positive given the excited tone. Laughter: Laughter often accompanies happy memories and a sense of lightheartedness. The speaker is reminiscing about someone and a specific moment, and the excitement suggests that the memory is a pleasant one. \\
\\
\hline
\textbf{\underline{Example 3}} \\
\textbf{Transcript:} \textit{\orange{You've got to tell me.}} \\
\textbf{Ground Truth Emotion Label:} \blue{Frustration} \\ \hdashline
\textbf{Label Only:} The speaker's speech suggesting a \blue{neutral} state. \\
\textbf{Interpretive Reasoning:} The speaker is \blue{frustrated} and wants the listener to tell them something. \\
\textbf{Evidence-Grounded Reasoning:} The speaker is \blue{frustrated}. They are likely feeling impatient, annoyed, or even angry because they feel like something is not being done or explained properly. The phrase \orange{"You've got to tell me"} suggests a sense of urgency and a demand for information. \\
\\
\hline
\end{tabular}
\caption{IEMOCAP: Qualitative comparison of emotion predictions. The first three samples in the evaluation dataset are presented. Each example includes the transcript, ground-truth emotion label, and model predictions with their inferred emotions. Emotion labels are in \blue{blue}, and quotes from the speech content are in \orange{orange}. Interpretive and evidence-grounded reasoning better align with the ground truth, offering richer justifications and improved emotion recognition.}
\label{table:audiollm responses iemocap}
\end{table*}

%% file: tables/audiollm_responses_meld_emotion.tex
\begin{table*}[t]
\centering
\small
\renewcommand{\arraystretch}{1.3}
\begin{tabular}{|p{15.5cm}|}
\hline
\textbf{\underline{Example 1}} \\
\textbf{Transcript:} \textit{\orange{Yeah, I'm not in that.}} \\
\textbf{Ground Truth Emotion Label:} \blue{Neutral} \\ \hdashline
\textbf{Label Only:} The speaker's emotion seems \blue{neutral}. \\
\textbf{Interpretive Reasoning:} The speaker is denying being in something, and they sound \blue{neutral} about it. \\
\textbf{Evidence-Grounded Reasoning:} The speaker's tone is \blue{neutral}, indicating they are not experiencing any strong emotions. They are simply stating a fact without any particular emphasis or feeling behind their words. \\
\\
\hline
\textbf{\underline{Example 2}} \\
\textbf{Transcript:} \textit{\orange{He's a dropper!}} \\
\textbf{Ground Truth Emotion Label:} \blue{Anger} \\ \hdashline
\textbf{Label Only:} The speaker's emotion seems \blue{neutral}.\\
\textbf{Interpretive Reasoning:} The speaker is expressing \blue{anger and frustration} towards someone they perceive as a \orange{"dropper,"} likely someone who is unreliable or lets them down. \\
\textbf{Evidence-Grounded Reasoning:} The speaker's tone is \blue{neutral}, but the statement itself carries a strong implication. While the speaker doesn't express anger or disgust directly, the phrase \orange{"He's a dropper!"} suggests a negative judgment about the person being referred to.  It implies that this person is unreliable, perhaps someone who makes promises they don't keep or fails to deliver on expectations. \\
\\
\hline
\textbf{\underline{Example 3}} \\
\textbf{Transcript:} \textit{\orange{Yes!}} \\
\textbf{Ground Truth Emotion Label:} \blue{Joy} \\ \hdashline
\textbf{Label Only:}  The speaker's speech reflects \blue{neutral}.\\
\textbf{Interpretive Reasoning:} The speaker said \orange{"Yes."} The emotion conveyed is \blue{neutral}. \\
\textbf{Evidence-Grounded Reasoning:} The speaker's tone is \blue{neutral}, indicating they are not experiencing any strong positive or negative emotions. They are likely calm, composed, and objective in their state of mind. \\
\\
\hline
\end{tabular}
\caption{MELD-ER: Qualitative comparison of emotion predictions. The first three samples in the evaluation dataset are presented. Each example includes the transcript, ground-truth emotion label, and model predictions with their inferred emotions. Emotion labels are in \blue{blue}, and quotes from the speech content are in \orange{orange}. Interpretive and evidence-grounded reasoning provide richer interpretation of speaker emotion by additionally making use of the speech content.}
\label{table:audiollm responses meld-er}
\end{table*}

%% file: tables/audiollm_responses_meld_sentiment.tex
\begin{table*}[t]
\centering
\small
\renewcommand{\arraystretch}{1.3}
\begin{tabular}{|p{15.5cm}|}
\hline
\textbf{\underline{Example 1}} \\
\textbf{Transcript:} \textit{\orange{Yeah, I'm not in that.}} \\
\textbf{Ground Truth Emotion Label:} \blue{Neutral} \\ \hdashline
\textbf{Label Only:} The speaker's sentiment is \blue{neutral}.\\
\textbf{Interpretive Reasoning:} The speaker is stating that they are not in a particular location or situation. They express this in a \blue{neutral} tone, indicating no strong feelings about the matter. \\
\textbf{Evidence-Grounded Reasoning:} The speaker's tone is \blue{neutral}, indicating they are not expressing strong positive or negative emotions. They are simply stating a fact without any noticeable inflection or emphasis. \\
\\
\hline
\textbf{\underline{Example 2}} \\
\textbf{Transcript:} \textit{\orange{He's a dropper!}} \\
\textbf{Ground Truth Emotion Label:} \blue{Negative} \\ \hdashline
\textbf{Label Only:} The speaker's sentiment is \blue{neutral}, based on their speech.\\
\textbf{Interpretive Reasoning:} The speaker is expressing a \blue{negative} sentiment towards someone they perceive as a \orange{"dropper,"} which likely implies a negative trait or action. \\
\textbf{Evidence-Grounded Reasoning:} The sentiment is \blue{negative}. The speaker's tone suggests disapproval or disgust towards the person being referred to as a \orange{"dropper."}  The word \orange{"dropper"} itself often carries a negative connotation, implying someone who is unreliable, messy, or in some way undesirable. \\
\\
\hline
\textbf{\underline{Example 3}} \\
\textbf{Transcript:} \textit{\orange{Yes!}} \\
\textbf{Ground Truth Emotion Label:} \blue{Positive} \\ \hdashline
\textbf{Label Only:} The speaker's sentiment is \blue{neutral}, based on their speech. \\
\textbf{Interpretive Reasoning:} The speaker said \orange{"Yes."} The sentiment is \blue{neutral}. \\
\textbf{Evidence-Grounded Reasoning:} The sentiment is \blue{positive}. The speaker's enthusiastic \orange{"Yes!"} conveys a strong sense of agreement, excitement, or approval. \\
\\
\hline
\end{tabular}
\caption{MELD-SR: Qualitative comparison of sentiment predictions. The first three samples in the evaluation dataset are presented. Each example includes the transcript, ground-truth sentiment label, and model predictions with their inferred sentiments. Sentiment labels are in \blue{blue}, and quotes from the speech content are in \orange{orange}. Interpretive and evidence-grounded reasoning better align with the ground truth, offering richer justifications and improved sentiment recognition.}
\label{table:audiollm responses meld-sr}
\end{table*}

%% file: custom.bbl
\begin{thebibliography}{47}
\providecommand{\natexlab}[1]{#1}

\bibitem[{Abouelenin et~al.(2025)Abouelenin, Ashfaq, Atkinson, Awadalla, Bach, Bao, Benhaim, Cai, Chaudhary, Chen, Chen, Chen, Chen, Chen, Chen, Chen, Dai, Dai, Fan, and Zhou}]{phi4}
Abdelrahman Abouelenin, Atabak Ashfaq, Adam Atkinson, Hany Awadalla, Nguyen Bach, Jianmin Bao, Alon Benhaim, Martin Cai, Vishrav Chaudhary, Congcong Chen, Dong Chen, Dongdong Chen, Junkun Chen, Weizhu Chen, Yen-Chun Chen, Yi-ling Chen, Qi~Dai, Xiyang Dai, Ruchao Fan, and Xiren Zhou. 2025.
\newblock {Phi-4-Mini} technical report: Compact yet powerful multimodal language models via mixture-of-{LoRAs}.
\newblock \emph{arXiv}.

\bibitem[{AI@Meta(2024)}]{llama3modelcard}
AI@Meta. 2024.
\newblock \href {https://github.com/meta-llama/llama3/blob/main/MODEL_CARD.md} {Llama 3 model card}.

\bibitem[{Ao et~al.(2024)Ao, Wang, Tian, Chen, Zhang, Lu, Wang, Li, and Wu}]{ao2024sdeval}
Junyi Ao, Yuancheng Wang, Xiaohai Tian, Dekun Chen, Jun Zhang, Lu~Lu, Yuxuan Wang, Haizhou Li, and Zhizheng Wu. 2024.
\newblock {SD-Eval}: A benchmark dataset for spoken dialogue understanding beyond words.
\newblock \emph{arXiv}.

\bibitem[{Bai et~al.(2023)Bai, Bai, Chu, Cui, Dang, and et~al.}]{bai2023qwen}
Jinze Bai, Shuai Bai, Yunfei Chu, Zeyu Cui, Kai Dang, and et~al. 2023.
\newblock Qwen technical report.
\newblock \emph{arXiv}.

\bibitem[{Bellver et~al.(2024)Bellver, Martín-Fernández, Bravo-Pacheco, Esteban, Fernández-Martínez, and D'Haro}]{bellver2024multimodal}
Jaime Bellver, Ivan Martín-Fernández, Jose Bravo-Pacheco, Sergio Esteban, Fernando Fernández-Martínez, and Luis D'Haro. 2024.
\newblock Multimodal audio-language model for speech emotion recognition.
\newblock In \emph{Odyssey: The Speaker and Language Recognition Workshop}.

\bibitem[{Busso et~al.(2008)Busso, Bulut, Lee, Kazemzadeh, Mower~Provost, Kim, Chang, Lee, and Narayanan}]{iemocap}
Carlos Busso, Murtaza Bulut, Chi-Chun Lee, Abe Kazemzadeh, Emily Mower~Provost, Samuel Kim, Jeannette Chang, Sungbok Lee, and Shrikanth Narayanan. 2008.
\newblock \href {https://doi.org/10.1007/s10579-008-9076-6} {{IEMOCAP}: Interactive emotional dyadic motion capture database}.
\newblock \emph{Language Resources and Evaluation}, 42:335--359.

\bibitem[{Chen et~al.(2022)Chen, Wang, Chen, Wu, Liu, and et~al.}]{chen2022wavlm}
Sanyuan Chen, Chengyi Wang, Zhengyang Chen, Yu~Wu, Shujie Liu, and et~al. 2022.
\newblock {WavLM}: Large-scale self-supervised pre-training for full stack speech processing.
\newblock \emph{IEEE Journal of Selected Topics in Signal Processing}, 16(6):1505--1518.

\bibitem[{Chen et~al.(2023)Chen, Wu, Wang, Liu, Tompkins, Chen, Che, Yu, and Wei}]{chen2023beats}
Sanyuan Chen, Yu~Wu, Chengyi Wang, Shujie Liu, Daniel Tompkins, Zhuo Chen, Wanxiang Che, Xiangzhan Yu, and Furu Wei. 2023.
\newblock {BEATs}: Audio pre-training with acoustic tokenizers.
\newblock In \emph{ICML}.

\bibitem[{Chiang et~al.(2023)Chiang, Li, Lin, Sheng, Wu, Zhang, Zheng, Zhuang, Zhuang, Gonzalez, Stoica, and Xing}]{vicuna2023}
Wei-Lin Chiang, Zhuohan Li, Zi~Lin, Ying Sheng, Zhanghao Wu, Hao Zhang, Lianmin Zheng, Siyuan Zhuang, Yonghao Zhuang, Joseph~E. Gonzalez, Ion Stoica, and Eric~P. Xing. 2023.
\newblock \href {https://lmsys.org/blog/2023-03-30-vicuna/} {Vicuna: An open-source chatbot impressing gpt-4 with 90\%* chatgpt quality}.

\bibitem[{Chu et~al.(2024)Chu, Xu, Yang, Wei, Wei, Guo, Leng, Lv, He, Lin, Zhou, and Zhou}]{Qwen2-Audio}
Yunfei Chu, Jin Xu, Qian Yang, Haojie Wei, Xipin Wei, Zhifang Guo, Yichong Leng, Yuanjun Lv, Jinzheng He, Junyang Lin, Chang Zhou, and Jingren Zhou. 2024.
\newblock {Qwen2-Audio} technical report.
\newblock \emph{arXiv}.

\bibitem[{Chu et~al.(2023)Chu, Xu, Zhou, Yang, Zhang, Yan, Zhou, and Zhou}]{chu2023qwenaudio}
Yunfei Chu, Jin Xu, Xiaohuan Zhou, Qian Yang, Shiliang Zhang, Zhijie Yan, Chang Zhou, and Jingren Zhou. 2023.
\newblock {Qwen-Audio}: Advancing universal audio understanding via unified large-scale audio-language models.
\newblock \emph{arXiv}.

\bibitem[{Das et~al.(2024)Das, Dingliwal, Ronanki, Paturi, Huang, Mathur, Yuan, Bekal, Niu, Jayanthi, Li, Mundnich, Sunkara, Srinivasan, Han, and Kirchhoff}]{Das2024SpeechVerse}
Nilaksh Das, Saket Dingliwal, S.~Ronanki, Rohit Paturi, David Huang, Prashant Mathur, Jie Yuan, Dhanush Bekal, Xing Niu, Sai~Muralidhar Jayanthi, Xilai Li, Karel Mundnich, Monica Sunkara, Sundararajan Srinivasan, Kyu~J Han, and Katrin Kirchhoff. 2024.
\newblock {SpeechVerse}: A large-scale generalizable audio language model.
\newblock \emph{arXiv}.

\bibitem[{D'efossez et~al.(2024)D'efossez, Mazar'e, Orsini, Royer, P'erez, J'egou, Grave, and Zeghidour}]{Defossez2024MoshiAS}
Alexandre D'efossez, Laurent Mazar'e, Manu Orsini, Am'elie Royer, Patrick P'erez, Herv'e J'egou, Edouard Grave, and Neil Zeghidour. 2024.
\newblock Moshi: a speech-text foundation model for real-time dialogue.
\newblock \emph{arXiv}.

\bibitem[{Deshmukh et~al.(2023)Deshmukh, Elizalde, Singh, and Wang}]{deshmukh2023pengi}
Soham Deshmukh, Benjamin Elizalde, Rita Singh, and Huaming Wang. 2023.
\newblock Pengi: An audio language model for audio tasks.
\newblock In \emph{NeurIPS}.

\bibitem[{Fu et~al.(2025)Fu, Wu, Wang, Zhang, Shan, Wu, and Liu}]{fu2025laerc}
Yumeng Fu, Junjie Wu, Zhongjie Wang, Meishan Zhang, Lili Shan, Yulin Wu, and Bingquan Liu. 2025.
\newblock {L}a{ERC}-{S}: Improving {LLM}-based emotion recognition in conversation with speaker characteristics.
\newblock In \emph{COLING}.

\bibitem[{Gemma~Team et~al.(2024)Gemma~Team, Pathak, Sessa, Hardin, Bhupatiraju, and et~al.}]{Riviere2024Gemma2I}
Morgane~Riviere Gemma~Team, Shreya Pathak, Pier~Giuseppe Sessa, Cassidy Hardin, Surya Bhupatiraju, and et~al. 2024.
\newblock Gemma 2: Improving open language models at a practical size.
\newblock \emph{arXiv}.

\bibitem[{Held et~al.(2024)Held, Li, Ryan, Shi, Zhang, and Yang}]{held2024diva}
William Held, Minzhi Li, Michael Ryan, Weiyan Shi, Yanzhe Zhang, and Diyi Yang. 2024.
\newblock Distilling an end-to-end voice assistant without instruction training data.
\newblock \emph{arXiv}.

\bibitem[{Hsu et~al.(2021)Hsu, Bolte, Tsai, Lakhotia, Salakhutdinov, and Mohamed}]{hsu2021hubert}
Wei-Ning Hsu, Benjamin Bolte, Yao-Hung~Hubert Tsai, Kushal Lakhotia, Ruslan Salakhutdinov, and Abdelrahman Mohamed. 2021.
\newblock {HuBERT}: Self-supervised speech representation learning by masked prediction of hidden units.
\newblock \emph{IEEE/ACM Trans. Audio, Speech and Lang. Proc.}, 29:3451–3460.

\bibitem[{Hu et~al.(2023)Hu, Zhou, Liu, Chen, Hao, Pan, Liu, Li, Sivasankaran, Liu, and Wei}]{hu2024wavllm}
Shujie Hu, Long Zhou, Shujie Liu, Sanyuan Chen, Hongkun Hao, Jing Pan, Xunying Liu, Jinyu Li, Sunit Sivasankaran, Linquan Liu, and Furu Wei. 2023.
\newblock {WavLLM}: Towards robust and adaptive speech large language model.
\newblock \emph{arXiv}.

\bibitem[{Ji et~al.(2024)Ji, Chen, Fang, Zuo, Lu, Wang, Jiang, Zhou, Liu, Cheng, Yang, Wang, Yang, Li, Jiang, He, Chu, Xu, and Zhao}]{ji2024wavchatsurvey}
Shengpeng Ji, Yifu Chen, Minghui Fang, Jialong Zuo, Jingyu Lu, Hanting Wang, Ziyue Jiang, Long Zhou, Shujie Liu, Xize Cheng, Xiaoda Yang, Zehan Wang, Qian Yang, Jian Li, Yidi Jiang, Jingzhen He, Yunfei Chu, Jin Xu, and Zhou Zhao. 2024.
\newblock {WavChat}: A survey of spoken dialogue models.
\newblock \emph{arXiv}.

\bibitem[{Kang et~al.(2024)Kang, Jia, Wu, Zhou, Lakomkin, Gaur, Sari, Kim, Li, Mahadeokar, and Kalinli}]{kang2024frozen}
Wonjune Kang, Junteng Jia, Chunyang Wu, Wei Zhou, Egor Lakomkin, Yashesh Gaur, Leda Sari, Suyoun Kim, Ke~Li, Jay Mahadeokar, and Ozlem Kalinli. 2024.
\newblock Frozen large language models can perceive paralinguistic aspects of speech.
\newblock \emph{arXiv}.

\bibitem[{Lee et~al.(2018)Lee, Wu, Liu, and Lee}]{lee2018spokensquad}
Chia-Hsuan Lee, Szu-Lin Wu, Chi-Liang Liu, and Hung-yi Lee. 2018.
\newblock {Spoken SQuAD}: A study of mitigating the impact of speech recognition errors on listening comprehension.
\newblock In \emph{Interspeech}.

\bibitem[{Li et~al.(2025{\natexlab{a}})Li, Liu, Dinkel, Niu, Zhang, and Luan}]{li2025rl}
Gang Li, Jizhong Liu, Heinrich Dinkel, Yadong Niu, Junbo Zhang, and Jian Luan. 2025{\natexlab{a}}.
\newblock Reinforcement learning outperforms supervised fine-tuning: A case study on audio question answering.
\newblock \emph{arXiv}.

\bibitem[{Li et~al.(2025{\natexlab{b}})Li, Liu, Dinkel, Niu, Zhang, and Luan}]{li2025r1aqa}
Gang Li, Jizhong Liu, Heinrich Dinkel, Yadong Niu, Junbo Zhang, and Jian Luan. 2025{\natexlab{b}}.
\newblock Reinforcement learning outperforms supervised fine-tuning: A case study on audio question answering.
\newblock \emph{arXiv}.

\bibitem[{Lu et~al.(2024{\natexlab{a}})Lu, Chen, Fu, Huang, Ginsburg, Wang, and yi~Lee}]{lu2024desta}
Ke-Han Lu, Zhehuai Chen, Szu-Wei Fu, He~Huang, Boris Ginsburg, Yu-Chiang~Frank Wang, and Hung yi~Lee. 2024{\natexlab{a}}.
\newblock {DeSTA}: Enhancing speech language models through descriptive speech-text alignment.
\newblock In \emph{Interspeech}.

\bibitem[{Lu et~al.(2024{\natexlab{b}})Lu, Chen, Fu, Yang, Balam, Ginsburg, Wang, and Lee}]{lu2025desta2}
Ke-Han Lu, Zhehuai Chen, Szu-Wei Fu, Chao-Han~Huck Yang, Jagadeesh Balam, Boris Ginsburg, Yu-Chiang~Frank Wang, and Hung-yi Lee. 2024{\natexlab{b}}.
\newblock {DeSTA2}: Developing instruction-following speech language model without speech instruction-tuning data.
\newblock In \emph{ICASSP}.

\bibitem[{Ma et~al.(2025)Ma, Chen, Wang, Chng, and Chen}]{ma2025audiocot}
Ziyang Ma, Zhuo Chen, Yuping Wang, Eng~Siong Chng, and Xie Chen. 2025.
\newblock {Audio-CoT}: Exploring chain-of-thought reasoning in large audio language model.
\newblock \emph{arXiv}.

\bibitem[{Ma et~al.(2024)Ma, Zheng, Ye, Li, Gao, Zhang, and Chen}]{emotion2vec}
Ziyang Ma, Zhisheng Zheng, Jiaxin Ye, Jinchao Li, Zhifu Gao, ShiLiang Zhang, and Xie Chen. 2024.
\newblock {emotion2vec}: Self-supervised pre-training for speech emotion representation.
\newblock In \emph{ACL}.

\bibitem[{{MERaLiON Team}(2024)}]{he2024meralionaudiollmtechnicalreport}
{MERaLiON Team}. 2024.
\newblock {MERaLiON-AudioLLM}: Bridging audio and language with large language models.
\newblock \emph{arXiv}.

\bibitem[{Ng et~al.(2025)Ng, Nguyen, Huang, Tai, Leong, Leong, Yong, Ngui, Susanto, Cheng, Rengarajan, Limkonchotiwat, Hulagadri, Teng, Tong, Siow, Teo, Lau, Tan, Ong, Ong, Montalan, Chan, Antonyrex, Lee, Choa, Tat-Wee, Liu, Tjhi, Cambria, and Teo}]{sealion}
Raymond Ng, Thanh~Ngan Nguyen, Yuli Huang, Ngee~Chia Tai, Wai~Yi Leong, Wei~Qi Leong, Xianbin Yong, Jian~Gang Ngui, Yosephine Susanto, Nicholas Cheng, Hamsawardhini Rengarajan, Peerat Limkonchotiwat, Adithya~Venkatadri Hulagadri, Kok~Wai Teng, Yeo~Yeow Tong, Bryan Siow, Wei~Yi Teo, Wayne Lau, Choon~Meng Tan, and 12 others. 2025.
\newblock {SEA-LION: Southeast Asian Languages in One Network}.
\newblock \emph{arXiv}.

\bibitem[{Panayotov et~al.(2015)Panayotov, Chen, Povey, and Khudanpur}]{panayotov2015librispeech}
Vassil Panayotov, Guoguo Chen, Daniel Povey, and Sanjeev Khudanpur. 2015.
\newblock Librispeech: An {ASR} corpus based on public domain audio books.
\newblock In \emph{ICASSP}.

\bibitem[{Poria et~al.(2019)Poria, Hazarika, Majumder, Naik, Cambria, and Mihalcea}]{poria2019meld}
Soujanya Poria, Devamanyu Hazarika, Navonil Majumder, Gautam Naik, Erik Cambria, and Rada Mihalcea. 2019.
\newblock {MELD}: A multimodal multi-party dataset for emotion recognition in conversations.
\newblock In \emph{ACL}.

\bibitem[{Radford et~al.(2023)Radford, Kim, Xu, Brockman, McLeavey, and Sutskever}]{radford2023whisper}
Alec Radford, Jong~Wook Kim, Tao Xu, Greg Brockman, Christine McLeavey, and Ilya Sutskever. 2023.
\newblock Robust speech recognition via large-scale weak supervision.
\newblock In \emph{ICML}.

\bibitem[{Sakshi et~al.(2025)Sakshi, Tyagi, Kumar, Seth, Selvakumar, Nieto, Duraiswami, Ghosh, and Manocha}]{sakshi2025mmau}
S~Sakshi, Utkarsh Tyagi, Sonal Kumar, Ashish Seth, Ramaneswaran Selvakumar, Oriol Nieto, Ramani Duraiswami, Sreyan Ghosh, and Dinesh Manocha. 2025.
\newblock {MMAU}: A massive multi-task audio understanding and reasoning benchmark.
\newblock In \emph{ICLR}.

\bibitem[{Shon et~al.(2023)Shon, Arora, Lin, Pasad, Wu, Sharma, Wu, yi~Lee, Livescu, and Watanabe}]{Shon2022SLUEPA}
Suwon Shon, Siddhant Arora, Chyi-Jiunn Lin, Ankita Pasad, Felix Wu, Roshan Sharma, Wei~Yu Wu, Hung yi~Lee, Karen Livescu, and Shinji Watanabe. 2023.
\newblock {SLUE Phase-2}: A benchmark suite of diverse spoken language understanding tasks.
\newblock \emph{ACL}.

\bibitem[{Sun et~al.(2024)Sun, Ahmed, Ma, Liu, Kabela, Pang, and Kalinli}]{sun2024contextual}
Chuanneng Sun, Zeeshan Ahmed, Yingyi Ma, Zhe Liu, Lucas Kabela, Yutong Pang, and Ozlem Kalinli. 2024.
\newblock Contextual biasing of named-entities with large language models.
\newblock In \emph{ICASSP}.

\bibitem[{Tang et~al.(2024)Tang, Yu, Sun, Chen, Tan, Li, Lu, Ma, and Zhang}]{tang2024salmonn}
Changli Tang, Wenyi Yu, Guangzhi Sun, Xianzhao Chen, Tian Tan, Wei Li, Lu~Lu, Zejun Ma, and Chao Zhang. 2024.
\newblock {SALMONN}: Towards generic hearing abilities for large language models.
\newblock In \emph{ICLR}.

\bibitem[{Touvron et~al.(2023)Touvron, Martin, Stone, Albert, Almahairi, and et~al.}]{Touvron2023Llama2}
Hugo Touvron, Louis Martin, Kevin~R. Stone, Peter Albert, Amjad Almahairi, and et~al. 2023.
\newblock Llama 2: Open foundation and fine-tuned chat models.
\newblock \emph{arXiv}.

\bibitem[{Wang et~al.(2024{\natexlab{a}})Wang, Zou, Lin, Sun, Liu, Zhang, Liu, Aw, and Chen}]{wang2024audiobench}
Bin Wang, Xunlong Zou, Geyu Lin, Shuo Sun, Zhuohan Liu, Wenyu Zhang, Zhengyuan Liu, AiTi Aw, and Nancy~F Chen. 2024{\natexlab{a}}.
\newblock {AudioBench}: A universal benchmark for audio large language models.
\newblock \emph{NAACL}.

\bibitem[{Wang et~al.(2025{\natexlab{a}})Wang, Zou, Sun, Zhang, He, Liu, Wei, Chen, and Aw}]{wang2025mnsc}
Bin Wang, Xunlong Zou, Shuo Sun, Wenyu Zhang, Yingxu He, Zhuohan Liu, Chengwei Wei, Nancy~F Chen, and AiTi Aw. 2025{\natexlab{a}}.
\newblock Advancing {Singlish} understanding: Bridging the gap with datasets and multimodal models.
\newblock \emph{arXiv}.

\bibitem[{Wang et~al.(2024{\natexlab{b}})Wang, Liao, Huang, Wu, Zong, and Zhang}]{wang2024blsp}
Chen Wang, Minpeng Liao, Zhongqiang Huang, Junhong Wu, Chengqing Zong, and Jiajun Zhang. 2024{\natexlab{b}}.
\newblock {BLSP-Emo}: Towards empathetic large speech-language models.
\newblock In \emph{EMNLP}.

\bibitem[{Wang et~al.(2025{\natexlab{b}})Wang, Sailor, Liu, and Aw}]{cpqa2025wang}
Qiongqiong Wang, Hardik Sailor, Tianchi Liu, and Ai~Ti Aw. 2025{\natexlab{b}}.
\newblock Contextual paralinguistic data creation for multi-modal speech-llm: Data condensation and spoken qa generation.
\newblock In \emph{Interspeech}.

\bibitem[{Wu et~al.(2024)Wu, Gong, Ai, Shi, Donbekci, and Hirschberg}]{Wu2024BeyondSL}
Zehui Wu, Ziwei Gong, Lin Ai, Pengyuan Shi, Kaan Donbekci, and Julia Hirschberg. 2024.
\newblock Beyond silent letters: Amplifying {LLMs} in emotion recognition with vocal nuances.
\newblock \emph{arXiv}.

\bibitem[{Xie et~al.(2025)Xie, Lin, Liu, Wu, Yan, and Miao}]{xie2025audioreasoner}
Zhifei Xie, Mingbao Lin, Zihang Liu, Pengcheng Wu, Shuicheng Yan, and Chunyan Miao. 2025.
\newblock {Audio-Reasoner}: Improving reasoning capability in large audio language models.
\newblock \emph{arXiv}.

\bibitem[{Xu et~al.(2024)Xu, Chen, Yu, Huang, Wu, Zhang, Li, Luo, and Gu}]{xu2024secap}
Yaoxun Xu, Hangting Chen, Jianwei Yu, Qiaochu Huang, Zhiyong Wu, Shi-Xiong Zhang, Guangzhi Li, Yi~Luo, and Rongzhi Gu. 2024.
\newblock {SECap}: speech emotion captioning with large language model.
\newblock In \emph{AAAI}.

\bibitem[{Zhao et~al.(2022)Zhao, Zhang, Hu, Liu, Jin, Wang, and Li}]{zhao-etal-2022-m3ed}
Jinming Zhao, Tenggan Zhang, Jingwen Hu, Yuchen Liu, Qin Jin, Xinchao Wang, and Haizhou Li. 2022.
\newblock {M}3{ED}: Multi-modal multi-scene multi-label emotional dialogue database.
\newblock In \emph{ACL}.

\bibitem[{Zhao et~al.(2025)Zhao, Zhu, Wang, Wang, Geng, Tian, and Xie}]{zhao2025steering}
Zhixian Zhao, Xinfa Zhu, Xinsheng Wang, Shuiyuan Wang, Xuelong Geng, Wenjie Tian, and Lei Xie. 2025.
\newblock Steering language model to stable speech emotion recognition via contextual perception and chain of thought.
\newblock \emph{arXiv}.

\end{thebibliography}
